# Neural Machine Translation on Scarce-Resource Condition: A case-study on Persian-English

Mohaddeseh Bastan     Shahram Khadivi[*]     Mohammad Mehdi Homayounpour

Computer Engineering and Information Technology Dept.
Amirkabir University of Technology, Tehran, Iran
Email: {m.bastan, khadivi, homayoun}@aut.ac.ir

*Abstract*— **Neural Machine Translation (NMT) is a new approach for Machine Translation (MT), and due to its success, it has absorbed the attention of many researchers in the field. In this paper, we study NMT model on Persian-English language pairs, to analyze the model and investigate the appropriateness of the model for scarce-resourced scenarios, the situation that exist for Persian-centered translation systems. We adjust the model for the Persian language and find the best parameters and hyper parameters for two tasks: translation and transliteration. We also apply some preprocessing task on the Persian dataset which yields to increase for about one point in terms of BLEU score. Also, we have modified the loss function to enhance the word alignment of the model. This new loss function yields a total of 1.87 point improvements in terms of BLEU score in the translation quality.**

## I. Introduction

Neural Networks are under great consideration. These networks have recently been used in many applications such as speech recognition [1], image processing [2], and natural language processing [3] and achieved remarkable results. Since the introduction of these networks and considerable results in different applications, many researchers in different fields are making use of the neural networks as a solution for their problems. MT which is a subcategory of natural language processing was firstly processed using neural networks by Castaño in 1997 [4].

For machine translation, these networks have been used for many different language pairs. In this paper, we propose a neural model for Persian translation for the first time. We use Tensorflow MT model [5] which was released by Google in 2015. We improve the base model with a new feature obtained from the statistical model. The new model consists of a new term as a cost function which measures the difference between the alignment obtained from neural model and statistical model. Then this cost is used to improve both accuracy and convergence time for the NMT.

The paper is organized as follow. In part II Statistical Machine Translation (SMT) and NMT and the corresponding mathematics are introduced. In part III literature review of NMT is done. In part IV our NMT model is presented. In part V the experiments and the improvements of the new model in comparison with the baselines are discussed. Finally, section VI concludes the paper

## II. Statistical and Neural Machine Translation

MT is the automation of the translation between human languages [6]. Two of the most successful models for machine translations are SMT and NMT which are discussed in light of the following subsections.

### A. Statistical Machine Translation

A common SMT model leads to find the target sentence *f*: $y_1, y_2, ..., y_T$ using source side sentence *e*: $x_1, x_2, ..., x_S$ by maximizing the following term [7]:

$$p(e|f) \sim p(e).p(f|e) \qquad (1)$$

In this equation, $p(e)$ is the language model which helps our output to be natural and grammatical, and *p(f|e)* is the translation model which ensures that *e* is normally interpreted as *f*, and not some other thing [8].

Most of the MT systems use log-linear model instead of the pure form, to model more features in the final equation. Then the model will be as follow [8]:

$$\log(p(e|f)) = \sum_{m=1}^{M} \lambda_m h_m(e.f) + logZ(e) \qquad (2)$$

This equation shows the m[th] feature of the SMT system with the $h_m$ symbol and the corresponding weight with the $\lambda_m$. The term *Z* is a normalization term which is independent from the weights. In Fig. 1 we see an architecture of an SMT. The model searches through different possibilities using its features as shown.

Alignment is one of the features for MT and the same alignment is used as what described in [10] for estimating parameters of SMT in this paper

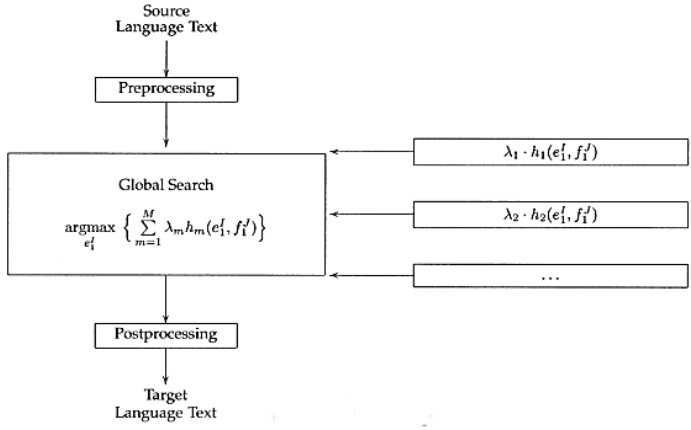

Figure 1. Architecture of Translation approach based on log-linear model [9]

*B. Neural Machine Translation*

Deep neural networks (DNNs) have shown impressive results in machine learning tasks. The success of these networks mostly is the result of the hierarchical aspects of these networks. DNNs are like pipeline processing in which each layer solves part of the issue and the result is fed into the next layer and at the end the last layer generates the output [11]. DNNs are powerful because the ability to perform parallel computations for several steps [12].

Most of the NMT models consist of two parts including an encoder which encodes the input sequence to a fixed length array, and a decoder which decodes the context vector into the output sequence [13]. Because the task is MT and the source and target sentences may have any lengths, the input and output of the NMT models are variable.

To address the following problem, recurrent neural networks (RNNs) are used for machine translation. RNNs are a map from feedforward neural networks into sequences. In each step $t$, the RNN computes the hidden state from the following equation, where $h_t$ is the hidden state at step $t$ and $x_t$ is the $t^{th}$ input in a sequence of inputs:

$$h_t = f(h_{t-1}.x_t) \quad (3)$$

$f$ is an activation function which can be simple as a sigmoid function or complicated as an LSTM [14]. Similarly, the next output symbol is computed using the following equation:

$$y_t = g(h_t) \quad (4)$$

Therefore, RNNs can easily map a sequence to another sequence. To map a sequence of input words to sequence of output words with different length, the first attempt is done by [13]. In this work, the input sequence is encoded to a fixed length context vector and the output sequence is generated by decoding the context vector. If $c$ is a context vector, the hidden state at the state $t$ is computed using the following equation:

$$h_t = f(h_{t-1}.y_{t-1}.c) \quad (5)$$

Each of the encoder and decoder is an RNN and the whole system is trained to maximize the following log-likelihood probability where N is the number of sentences in training set, $Y^n$ is the target output corresponding to source input $X^n$:

$$\max_\theta (\frac{1}{N} \sum_{n=1}^{N} \log P_\theta(Y^n|X^n)) \quad (6)$$

The above model works fine for small sentences. But as the length of the sentence increases, the context vector cannot encode all of the source sentences and the performance decreases significantly [15]. So, the context vector is a bottleneck for this model and a vector with fixed length should be revised. Paper [16] proposes a model which does not encode the whole source sentence into a fixed length array. Instead, the input sentence is encoded to a sequence of arrays and a subset of these arrays are selected for decoding. Then, the model can translate the longer sentences easily. In the new model, each conditional probability is defined as follows:

$$p(y_i|y_1.y_2.\ldots.y_{i-1}.X) = g(y_{i-1}.s_i.c_i) \quad (7)$$

Where $y_i$ is the $i^{th}$ word of the output and $X$ is the input sentence, the $s_i$ is the hidden network state at the $i^{th}$ step and is computed in this way:

$$s_i = f(s_{i-1}.y_{i-1}.c_i) \quad (8)$$

In spite of the conventional encoder-decoder method, in this equation for each output $y_i$ the probability is conditional on corresponding $c_i$. Each $c_i$ is computed as weighted sum of the annotations $h_i$ as follows:

$$c_i = \sum_{j=1}^{T_x} \alpha_{ij} h_j \quad (9)$$

In this equation, $T_x$ is the length of the source sentence and $\alpha_{ij}$ is the weight for the $j^{th}$ annotation and is computed as follow:

$$\alpha_{ij} = \frac{\exp(e_{ij})}{\sum_{k=1}^{T_x} \exp(e_{ik})} \quad (10)$$

Finally, $e_{ij}$ is the alignment model which shows how the words around the input position $j$ are compliance with the $i^{th}$ output position. The alignment model is a feedforward neural network which is trained simultaneously with the other components of the network. In contrast of the other NMT, the alignment is not a hidden variable here and is computed as a soft alignment [16].

For training the model we use Stochastic Gradient Descent [17]. The learning rate is a parameter which controls how large a step should be in the direction of the negative gradient [18]. It is controlled adaptively here. If the improvement in terms of the loss function is not seen over last three iterations, the model will decay the learning rate by a specific factor.

We take advantage of this soft alignment and use it to train the model faster and also more accurate. We also trained the model for Persian language for the first time and adjusted the parameters and hyper parameters. Finally, we added a feature from statistical model to make the soft alignment more powerful which results in the decrease in convergence time and improves the model. we added another term to the conditional decay which described above. In implementation we are heavily relying on the Tensorflow translation model.

### III. RELATED WORK

In 2003 neural network language models generally introduced in [19]. In machine translation, some researchers used these models for rescoring the translation [20]. For example, [12] used neural networks for rescoring the translation candida sentences and [13] used neural networks for translation scores in phrase table. One of the simplest and most impressive works in NMT is [21] which used neural networks for rescoring the n-best list in an MT system. This improved MT effectively. In 2012, Li proposed an MT model using feedforward neural networks which used an output layer for classification and a short list for rescoring [22].

For language model and machine translation, it has been shown that RNNs empirically work better than feedforward neural networks [23]. Most of the RNN translation models are as an encoder-decoder family. encoder-decoder models for MT first used in [24]. They used a convolutional neural network (CNN) to encode the input sentences into an array and then used an RNN for decoding.

A newer model for encoder-decoder was presented in [25], where the decoder was conditioned on the source sentences. In this work a language model is combined with a topic model and the results show some improvements in rescoring. In [13] another encoder-decoder was introduced which used an LSTM for encoding and decoding. The authors mostly considered on combining their neural network with an SMT model.

NMT models have two main problems which researchers are trying to solve. First, the ability of the model to translate decreases as the length of sentences increase. Bahdanau used an attention model to address the problem of translating long sentences [16]. Next is the memory issue. As the size of the corpus increases, the accuracy of the translation increases, but the problem of memory usage emerges. In [26] a model was proposed to address this problem. They tried to translate some part of the input sentences like phrase based translation.

In [27] a model for scoring phrases was proposed. In this model, a feedforward neural network with input and output of fixed length is used. Devlin [28] proposed an NMT model using feedforward neural network. In his model a language model encoder using neural network and a decoder from MT model combined and used decoder alignment information for the language model to output the most useful words corresponding to the input sentences. This model made a significant improvement in machine translation, but the limitation of the length of the sentences yet remained.

Bidirectional LSTM first proposed by [29] and used for speech recognition task. These networks were used for MT in [30] and created a strong model which used next and previous input words for translation. The idea of guided alignment first proposed in [31]. And our proposed model for using both SMT and NMT alignments is inspired by this paper.

### IV. PERSIAN NEURAL TRANSLATION MODEL

In this section we define the issue for Persian translation and the preprocessing needs to be done before feeding the input into the model. Then the proposed model for NMT which uses soft alignment and SMT alignment feature for translation is described.

#### C. Data preprocessing

The Persian language makes MT a difficult task because of its specific characteristics. So the input sentences should be preprocessed and then fed into the NMT model. Here is the list of preprocessing tasks which are done on Persian corpora:

- All corpora are changed to have one sentence per line ending in one of the punctuations: '؟', '.', or '!'.
- All words are separated with a single space symbol.
- All zero-width non-breakings have been removed. For instance, the word "می‌نویسم" is changed to "می نویسم"
- All the adherent words have been tokenized. For instance, the word "آنها" is changed to "آن ها"
- If a word is an adherent with a symbol, punctuation sign or other characters, it is disparted.

All of these preprocessing tasks, prepare the Persian data to be used for NMT. The first two preprocessing tasks in the above list, are general which should be done for every language pairs and every MT models. The next two are Persian specific. Unlike the SMT, for NMT we use these two preprocessing to distinguish the words. This is a tradeoff between the number of unique words and the length of the sentences. As the problem of the length of the sentence is decreased after using techniques described in [16], we decided to decrease the number of unique words and increase the length of the sentences. This configuration leads to better results. The last one results into disjointing non-related characters. If we do not dispart the word and its adjunct punctuation sign, the system considers them as a whole word, and this is not acceptable for us. Since the NMT system should get them as two distinct words and not one word.

#### D. The alignment feature

One of the properties of the NMT models is that it doesn't need to define different features and each feature is tuned to maximize the probability function. Indeed, it learns everything via a unique model and translates the source sentence into the target via the trained model. On the other hand, the SMT defines different features and computes the corresponding weights for each of them and tries to maximize the probability function. Because each of them has its own pros. Our model benefits both of these features and tries to increase the accuracy of the alignment model in NMT using alignment model in SMT.

In SMT model, we use the GIZA++ [32] tool to align the source and target sentences to each other. This tool uses an EM algorithm to align words in source and target sentences and shows which words of the source sentence is aligned with which word or words of the target sentence. This alignment can be defined as the following matrix:

$$M^{T \times S} : \begin{cases} M[i,j] = 1 & \text{If the i}^{th} \text{ target word is aligned to the j}^{th} \text{ source word} \\ M[i,j] = 0 & \text{Otherwise} \end{cases} \quad (11)$$

Here $M$ is the alignment matrix, $S$ is the size of the source sentence and $T$ is the size of the target sentence. We name this matrix as EM-alignment matrix.

In NMT model, we use the soft alignment. As described in part II-B it makes a matrix for each step of the training phase. This matrix is the $e$ matrix in (10). The i$^{th}$ row and j$^{th}$ column of this matrix defines the compliance of the i$^{th}$ target word with the j$^{th}$ source word. So we have another matrix which is the output of each step of the NMT model. We name this matrix as NMT-alignment matrix

What we do here is adding a cost function to NMT model consisting of the difference between these two matrices. The fact behind it, is that a fully translated model using EM-alignment definitely has a better alignment model than an NMT model which is in the process of training and is not convergent. This helps the NMT model to converge faster and at the same time find the alignment model which fits to the corpora. Here is the term which is added to previous cost function of NMT:

$$\frac{\omega}{2(|T|+|S|)} f\left(\left|M^{T \times S} - e^{T \times S}\right|\right) \quad (12)$$

In this equation, $e$ is the NMT-alignment matrix and $M$ is the EM-alignment matrix. The $|\ |$ symbol means the absolute value. Function $f$, is the summation of all elements of the matrix. The term before the function is for normalizing the summation. And $\omega$ is a weight which defines how important this term is versus default cost function. The higher the $\omega$, the more important the alignment difference is. We arbitrary set this weight to 0.2. This weight seems reasonable, because the cost function itself consists of the difference between the model translation and the target translation and this difference is more important than the alignment difference.

The cost function in this model is used for learning rate decay factor. In an NMT at each step we expect the model to decrease the cost function. If after series of iterations, the cost function did not decrease, the learning rate will be changed.

Adding a new term to cost function helps the model to learn the alignment more accurately. If the NMT-alignment have a wide margin with EM-alignment, the model will be penalized. So it will learn to align the source and target sentences with more attention to EM-alignment. At the end, we expect the model to have an alignment closer to EM-alignment, unless the alignment is in contrast with translation. Since in cost function the translation is weightier, the model will not suffer from wrong EM-alignments and it won't change the correct NMT-alignments.

## V. EXPERIMENTS

In this section the experiments done for the proposed model with the results and analysis of each experiment are described. First, system configurations, then the datasets and finally, the experiments and results are described.

### E. System configurations

For our experiment, we use NVIDIA GeForce GTX 780 GPU which increases the speed of processing in comparison with CPU. Also NVIDIA CUDA toolkit v.7 is used specifically for its math libraries and optimization routines. Also we take advantageous of cuDNN library v5 for increasing the training speed. For programming, we used Tensroflow framework v0.10 and made our changes based on MT model proposed by the providers.

### F. Dataset description

We used two datasets. The first dataset is Verbmobil English-Persian translation dataset which consists of some conversation sentences in English and their translations in Persian. The second dataset is a transliteration dataset. It consists of some separated characters of words in Persian. In this dataset, the sentence means a word with separated characters, and words mean characters. In Table I the information about Training, Development and Test sets of Verbmobil and Transliteration are provided. **Sent. Count** means the number of sentences for each of the English and Persian datasets. **Unique words count** is the number of distinct words available in each corpus. This number is for the main dataset without any preprocessing task. In Table II, an example from each dataset is described. One sentence for each dataset is shown in Table II.

### G. Evaluating Measurements

For evaluating the proposed model, we use different measurements. For translating we use BLEU [33] measurement, which is quick and language independent. This measure is based on precision of n-grams of the translated text in comparison with target reference or references.

For transliteration task we use four different measures in addition to BLEU. The first measurement is the accuracy. Accuracy means how many sentences have been transliterated completely without any error in any position of the sentence. The next measurement is WER (Word Error Rate), which counts the number of words transliterated incorrectly. The words should be transliterated exactly in the same order as source words. As we described earlier, words here mean the character. So WER measures the number of characters which has been transliterated into the wrong character.

TABLE I. DATASETS DESCRIPTION

| Dataset | Training | | | | Development | | | | Test | | | |
|---|---|---|---|---|---|---|---|---|---|---|---|---|
| | Persian | | English | | Persian | | English | | Persian | | English | |
| | Sent. Count | Unique Words Count | Sent. Count | Unique Words Count | Sent. Count | Unique Words Count | Sent. Count | Unique Words Count | Sent. Count | Unique Words Count | Sent. Count | Unique Words Count |
| Verbmobil | 26142 | 5909 | 26142 | 3118 | 276 | 463 | 276 | 350 | 250 | 429 | 250 | 345 |
| Transliteration | 88507 | 35 | 88507 | 54 | 1000 | 27 | 1000 | 26 | 1000 | 26 | 1000 | 26 |

TABLE II. DATASET EXAMPLE

| Dataset | English | Persian |
|---|---|---|
| Verbmobil | Monday the eighth of November would suit me fine | . دوشنبه هشتم نوامبر برای من خوب است |
| Transliteration | A A m i n | آ م ی ن |

The third measure is PER (Position-independent word Error Rate) [34]. It is the same as WER but ignoring the order of the words. This measure looks at the sentences as bag-of-the-words and does not consider the position of the words. Then it counts the number of words translated incorrectly and are not in the target sentence at all. Finally, the last measure is TER (Translation Error Rate) [35] which counts the number of edits required to change a system output into one of the given translation references.

*H. Experiments and Results*

We evaluate our model by 3 different experiments. First, we find the best configuration for each dataset. The parameters adjusted are number of layers and number of nodes in each layer of RNN. After adjusting the parameters and hyper parameters for each dataset, we evaluate our proposed model. First, using the best adjusted model, the changes to dataset and then the cost function effect is evaluated. The results are shown in Table III through VI. First we describe the results on transliteration task and then for the translation task.

Table III shows different configurations of the NMT model for transliteration task. As we see by increasing the number of hidden layer nodes, all measurements improve (BLEU and accuracy increase and TER, WER, PER decrease). But stops at a specific configuration, where increasing the number of hidden nodes does not improve the model[1]. For transliteration task, we do not have any preprocessing step, since all the words are separated by space and there is no punctuation mark or any symbol except the default Persian and English alphabet characters. So the next experiment is adding new cost function to the default cost function. The results are shown in Table IV. Changing cost function improves the model significantly. That is mostly because the previous cost function does not include the EM-alignment and suffers from aligning incorrectly.

Next experiments are on Verbmobil dataset and translation task. For this dataset first we configure the model for best parameters and hyper parameters. The results are shown in Table V. As we expect, by increasing the number of hidden nodes the model works better, because the task is translation and the number of unique words are more than the number of unique words in transliteration task, the model responses better by increasing the number of hidden nodes. In Table VI the preprocessing task and cost function are added to the best baseline. As it can be seen due to preprocessing, the BLEU measure increases for about one unit which shows the importance of preprocessing. The new cost function increases base-line system for about 1.87 units in BLUE measure which shows that the new cost function also works effectively for translation task.

TABLE III. TRANSLITERATION CONFIGURATION RESULTS

| Number of layers | Number of hidden nodes | BLEU (%) | Accuracy (%) | TER (%) | WER (%) | PER (%) |
|---|---|---|---|---|---|---|
| 3 | 50 | 74.04 | 50.3 | 12.27 | 13.12 | 8.30 |
| 4 | 50 | 72.8 | 55 | 12.34 | 12.55 | 8.28 |
| 3 | 100 | 73.87 | 50.7 | 12.22 | 12.32 | 8.34 |
| 4 | 100 | 76.21 | 44.7 | 11.04 | 11.12 | 8.00 |
| 3 | 200 | 68.97 | 48.9 | 14.04 | 14.61 | 9.31 |
| 4 | 200 | 75.31 | 47.1 | 12.15 | 12.17 | 8.12 |

TABLE IV. TRANSLITERATION COST FUNCTION RESULTS

| Cost function | BLEU | Accuracy | TER | WER | PER |
|---|---|---|---|---|---|
| With EM-alignment | 76.21 | 44.7 | 11.04 | 11.12 | 8.00 |
| Without EM-alignment | 77.13 | 44.2 | 11.01 | 11.02 | 7.77 |

TABLE V. TRANSLATION CONFIGURATION RESULTS FOR

| Number of layers | Number of hidden nodes | BLEU (En-Fa) | BLEU (Fa-En) |
|---|---|---|---|
| 3 | 500 | 16.21 | 20.10 |
| 4 | 500 | 16.5 | 20.25 |
| 3 | 1000 | 18.15 | 21.69 |
| 4 | 1000 | **18.33** | **21.88** |
| 3 | 2000 | 17.92 | 21.25 |
| 4 | 2000 | 18.12 | 21.5 |

TABLE VI. VERBMOBIL PREPROCESSING AND COST FUNCTION RESULTS

| Cost function | BLEU (En-Fa) | BLEU (Fa-En) |
|---|---|---|
| Base line | 18.33 | 21.88 |
| + preprocessing | 19.25 | 22.80 |
| + preprocessing + new cost function | 19.75 | 23.65 |

---
[1] Experiments consist more adjustment for the model, the bests are reported.

VI. CONCLUSION

In this paper the first NMT system for Persian language which is trained using scarce data was proposed. The parameters and hyper parameters of the model are adjusted for Persian to English language. Also some preprocessing were tasks introduced which help the Persian model to be translated accurately. Finally, a cost function was added to soft-alignment in neural machine translations. The whole system increased the performance of the base-line system for about 1.87 for translation task and about 0.9 for transliteration task.